\documentclass[conference, titlepage]{IEEEtran}
\IEEEoverridecommandlockouts

\usepackage[style=ieee,sorting=none,backend=bibtex]{biblatex}
\usepackage{amsmath}
\usepackage[ruled,vlined]{algorithm2e}

\usepackage{lipsum}
\usepackage{wrapfig}
\usepackage{subfig}
\usepackage[pdftex]{graphicx}
\graphicspath{{./images/}}

\usepackage[dvipsnames]{xcolor}

\DontPrintSemicolon

\begin{document}

\title{Abstract Demonstrations and Adaptive Exploration for Efficient and Stable Multi-step Sparse Reward Reinforcement Learning\\
\thanks{Xintong Yang thanks the Chinese Scholarship Council (CSC) for providing the living stipend for his Ph.D. programme (No. 201908440400).}
}

\author{\IEEEauthorblockN{Xintong Yang, Ze Ji}
\IEEEauthorblockA{\textit{School of Engineering} \\
\textit{Cardiff University}\\
Cardiff, UK \\
yangx66@cardiff.ac.uk, jiz1@cardiff.ac.uk}
\and
\IEEEauthorblockN{Jing Wu, Yu-Kun Lai}
\IEEEauthorblockA{\textit{School of Computer Science and Informatics} \\
\textit{Cardiff University}\\
Cardiff, UK \\
wuj11@cardiff.ac.uk, laiy4@cardiff.ac.uk}
}

\maketitle

%===============================================================================

\begin{abstract}
    Although Deep Reinforcement Learning (DRL) has been popular in many disciplines including robotics, state-of-the-art DRL algorithms still struggle to learn long-horizon, multi-step and sparse reward tasks, such as stacking several blocks given only a task-completion reward signal. To improve learning efficiency for such tasks, this paper proposes a DRL exploration technique, termed $\mathbf{A^2}$,  which integrates two components inspired by human experiences: Abstract demonstrations and Adaptive exploration. $\mathbf{A^2}$ starts by decomposing a complex task into subtasks, and then provides the correct orders of subtasks to learn. During training, the agent explores the environment adaptively, acting more deterministically for well-mastered subtasks and more stochastically for ill-learnt subtasks. Ablation and comparative experiments are conducted on several grid-world tasks and three robotic manipulation tasks. We demonstrate that $\mathbf{A^2}$ can aid popular DRL algorithms (DQN, DDPG, and SAC) to learn more efficiently and stably in these environments.
\end{abstract}

%===============================================================================

\section{INTRODUCTION}

\begin{figure}[t]
	%\vspace{-15pt}
    \centering
	\includegraphics[width=0.48\textwidth]{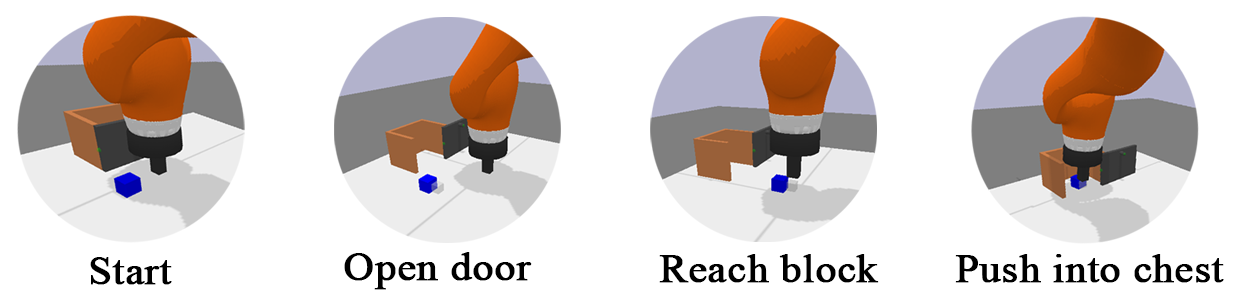}
	\caption{Visualisation of a pushing task.
	\label{fig:pushing}}
	%\vspace{-15pt}
\end{figure}

Deep Reinforcement Learning (DRL) has achieved exciting advances recently in robotics \cite{lazaridis2020deep}. However, DRL agents still struggle to solve many robotic tasks with long horizon, multiple steps and sparse rewards. For example, in Fig.~\ref{fig:pushing}, a robot is asked to push a block into a closed chest, where the chest needs to be opened before the block can be pushed into it, and only a reward is given when the task is done. Fortunately, humans provide many useful insights for mastering such tasks. In this paper, we revisit and integrate two ideas inspired from human experiences to make DRL more efficient and stable in these difficult tasks.

In the real world, humans benefit substantially from decomposing a complex task into a sequence of subtasks. This helps divide the task into a number of steps that are easier and faster to solve \cite{dantam2016incremental, levine2012monitoring, cobo2012automatic}. We can then accomplish the complex task by achieving each step in a specific order, such as following an instruction manual to assemble different parts of a piece of furniture. In addition, skills developed during learning one step can be reused in other steps, such as skills to stack a specific block can be reused to stack others. The example in Fig.~\ref{fig:pushing} shows that the three steps of the pushing task are easier to achieve individually and they share similar arm motions (reaching and pushing).

Inspired by this, we propose firstly to leverage \textbf{abstract demonstrations}, which provide the correct order of steps to a learning agent instead of low-level motions. It has been shown that, given an efficient enough algorithm to learn the subtasks such as hindsight experience replay, abstract demonstrations can accelerate learning for multi-step tasks \cite{yang2021hierarchical}. The main benefits of abstract demonstrations when compared to demonstrations of motion trajectories \cite{hussein2017imitation, ravichandar2020recent} are: 1) they do not encode a specific pattern of the behaviours when solving a task; 2) they release robotic operators from the tedious processes of collecting motion trajectory data.

The second idea is inspired from humans' \textbf{adaptive exploration} strategies. For example, children do not try to pick and stack Lego blocks randomly when they have already known which blocks they need and how to stack them together. This relates closely to the trade off between exploration and exploitation, a core issue in DRL \cite{sutton2018reinforcement}. Although exploration methods have been proposed for long-horizon tasks \cite{nair2018overcoming}, its adaptivity has rarely been studied in the context of multi-step tasks. In multi-step tasks, the learning progress of later steps depends heavily on former steps. For example, stacking the fourth block requires the previous three blocks to have been stacked well. As a result, an agent that explores constantly or decays exploration in a task-agnostic way, which is the default setting in most works \cite{wang2020deep}, will have difficulty approaching later steps. With this in mind, we propose to  adjust exploration adaptively based on an agent's performance on each of the task steps.

Previously, these ideas have been explored individually (see a detailed review in Section~\ref{sec:related-work}). In this paper, we highlight the effectiveness of their integration into DRL algorithms for learning multi-step, sparse reward tasks. We conducted a series of ablation studies on three representative off-policy DRL agents, i.e., DQN (deterministic, discrete actions) \cite{mnih2015dqn}, DDPG (deterministic, continuous actions) \cite{lillicrap2015continuous} and SAC (stochastic, continuous actions) \cite{haarnoja2018soft}, using the Mini-grid \cite{gym_minigrid} and Pybullet Multigoal environments \cite{yang2021open}. 

The results show that \textbf{abstract demonstrations} can accelerate learning and improve performance slightly in the grid-world tasks (DQN) and significantly in continuous control tasks (DDPG and SAC). \textbf{Adaptive exploration} does not further improve task success rates in general, but it does stablise learning with reduced performance variances. In short, our contributions include:

\begin{itemize}
\item[$\bullet$] We propose $\mathbf{A^2}$, a method which integrates \textbf{A}bstract demonstrations and \textbf{A}daptive exploration, for more efficient and stable multi-step sparse reward DRL.

\item[$\bullet$] We discuss possible implementations of the adaptive exploration component of $\mathbf{A^2}$ for modern DRL algorithms, including discrete and continuous actions, deterministic and stochastic policies.

\item[$\bullet$] We validate the effectiveness of $\mathbf{A^2}$ in various settings: with a deterministic DQN agent (discrete actions, grid-world tasks), a deterministic DDPG agent (continuous actions, robotic tasks) and a stochastic SAC agent (continuous actions, robotic tasks).

\end{itemize}

\textbf{Paper organisation:} Section~\ref{sec:related-work} draws connections between $\mathbf{A^2}$ and existing works. Section~\ref{sec:preliminaries} briefly reviews important preliminaries for using $\mathbf{A^2}$. Section~\ref{sec:methods} describes $\mathbf{A^2}$ in details. Section~\ref{sec:result} presents experimental results and discussions. Section~\ref{sec:conclusion} concludes the paper.

%===============================================================================

\section{Related work}
\label{sec:related-work}

\textbf{Long-horizon multi-step tasks} have been addressed by task and motion planning (TAMP) \cite{dantam2016incremental}. TAMP typically generates high-level task sequences and plans motions for each subtask individually \cite{levine2012monitoring, fox2003pddl2}. Although the idea of task decomposition based on human priors has been around for decades \cite{fikes1971strips}, researchers have just started recently to study how it can help learning-based methods \cite{jurgenson2020sub, andreas2017modular}. Many works in RL focus on learning sub-policies (or options) that are specific to different subtasks or manifest different skills \cite{Sutton1999, krishnan2017ddco}. Our work differs from them in that we decompose a task into subtasks to guide the learning of a control policy (as demonstrations), while they require an agent to learn to decompose a task during learning. In principle, our method can be incorporated with any task decomposition method, be it based on human priors or learning \cite{shiarlis18a}.

\textbf{Learning from demonstrations (LfD)} is a practical approach to teach robots behaviours using trajectories collected by another agent, typically a human. These behaviours are normally difficult to program \cite{ravichandar2020recent}. DRL also benefits from demonstrations in sparse reward and/or long-horizon tasks (also known as imitation learning) \cite{hussein2017imitation}. Different from the mainstream works, which use trajectories at the control command level \cite{nair2018overcoming}, we use an abstract form of demonstrations, i.e., the correct sequence in which task steps should be performed and learnt. This reduces the human labour spent on collecting long robot motion trajectories. For cases where some subtasks are too difficult to learn, motion trajectories can still be used to further accelerate learning for the subtasks. In order words, abstract demonstrations could be an alternative or a supplement to motion trajectories demonstrations.

\textbf{Automatic curriculum learning (ACL}) has become increasingly active in the field of RL and robotics \cite{acl2020portelas}. Briefly, ACL methods automatically adapt the distribution of training data for a learning agent \cite{acl2020portelas}. Our method can be seen as a contribution to ACL, which 1) prompts the agent to learn from easier subtasks to harder subtasks in the correct sequences (abstract demonstrations) and 2) adjusts the agent's exploration behaviours based on its performance on each subtask.

%===============================================================================

\section{Preliminaries}
\label{sec:preliminaries}

\textbf{Markov Decision Process} is a tuple $\langle S, A, R, p, \gamma, \rho_0 \rangle$, where $S$ is the state space, $A$ the action space, $R(s,a)$ the reward function, $p(s'|s,a)$ the system transition model, $\gamma$ the discount factor and $\rho_0$ the initial state distribution. A policy $\pi(a|s)$ is a mapping from states to actions. A state-action (Q) value function, $Q^{\pi}(s_0, a_0)$, is defined as the expected, discounted and accumulated rewards starting from taking an action $a$ at state $s$ and following a policy $\pi$ thereafter, i.e., $Q^{\pi}(s_0, a_0)=\mathrm{E}_{a\sim\pi,s\sim p}[\sum_{t=0}^{T}\gamma^{t}R(s_t,a_t)]$, where $T$ is the maximal number of task timesteps. The goal of an RL algorithm is to find an optimal policy that maximises the value function \cite{sutton2018reinforcement}.

\textbf{Goal-conditioned Reinforcement Learning}: The GRL problem operates on an MDP augmented with a goal space $G$ \cite{Andrychowicz2017}. Instead of pursuing a single goal (conveyed by a single-objective reward function) as in standard RL, a GRL agent seeks to maximise a universal value function, $Q^{\pi}(s,g,a)$. Commonly, a goal is defined as some kind of transformation of a state, $g=m(s)$, assuming that, at any state, there is always a goal that is achieved if the agent arrives at that state.

\textbf{Deep Q Network}: DQN is an off-policy, deterministic RL algorithm that uses neural networks to approximate the Q function for discrete action tasks \cite{mnih2015dqn}. Commonly, DQN uses the \textit{epsilon-greedy} (EGr) method for exploration. It takes a random action with a probability $\epsilon$ and takes an action according to the learnt Q function with a probability $1-\epsilon$.

\textbf{Deep Deterministic Policy Gradient}: DDPG is an off-policy actor-critic style DRL algorithm developed for continuous control tasks \cite{lillicrap2015continuous}. It approximates a Q-function and a deterministic policy using separate neural networks, alternating the updates of the two with some interval. Originally, Ornstein–Uhlenbeck noises are added to the learnt policy to form exploratory actions \cite{lillicrap2015continuous}. Recently, researchers tend to use a simpler, but efficient exploration strategy, which takes random actions with a probability of $\epsilon$ and takes learnt actions with Gaussian noises with a probability of $1-\epsilon$ \cite{Andrychowicz2017}. For convenient reading, we name this strategy \textit{epsilon-Gaussion} (EGa).

\textbf{Soft Actor Critic}: SAC is an off-policy, stochastic DRL algorithm \cite{haarnoja2018soft}. It updates the policy and value network with entropy regularisation. Typically, the policy network outputs the mean and standard deviation for a Gaussian policy. SAC controls its exploration by adjusting the temperature parameter $\alpha$ either manually or automatically. One can also alter the standard deviation of the policy to directly control exploration in a heuristic manner.

\textbf{Hindsight Experience Replay}: HER is a goal-relabelling technique for GRL, which copies transitions and replaces their desired goals with some other goals obtained by a sampling strategy \cite{Andrychowicz2017}. It improves learning significantly in goal-conditioned sparse reward tasks. Originally, Andrychowicz et al. proposed in \cite{Andrychowicz2017} the \texttt{final}, \texttt{random}, \texttt{episode} and \texttt{future} strategies to sample goals for replacement, and demonstrated that the \texttt{future} strategy is the most efficient. It copies a transition $k$ times, samples $k$ future transitions, and replaces the desired goals for each copied transition with the sampled ones. We adopt the \texttt{future} strategy in all our experiments, with $k=4$.

%===============================================================================

\section{Methods}
\label{sec:methods}

This section illustrates our proposed $\mathbf{A^2}$ method in details. Overall, it includes two components. First, abstract demonstrations are provided to guide the robot to learn subtasks in the correct order, leading to the completion of a task. Secondly, an adaptive exploration technique adjusts the exploration parameters to achieve faster and more stable learning.

\subsection{Abstract demonstrations}

We propose to first decompose a multi-step task into a sequence of subtasks that are easier to accomplish. We observed that, in many tasks (e.g., stacking blocks), although the decomposed steps may have different purposes (e.g., stacking each of the blocks), the underlying behaviours achieving them often share common characteristics (e.g., move-pick-move-place). This inspired us to use goal-conditioned reinforcement learning (GRL) \cite{Andrychowicz2017} as our main learning framework. The reason is that GRL was proved to have the potential to learn shared representations and knowledge between the steps \cite{jurgenson2020sub}, and Hindsight Experience Replay (HER) enables a GRL agent to learn short-term goals efficiently with sparse reward signals \cite{Andrychowicz2017}.

The motivation of using abstract demonstrations over demonstrations at the control command level (kinematic demonstrations) is twofold. First, abstract demonstrations do not encode a specific pattern of behaviours for solving a subtask. Second, it releases humans from the tedious processes of collecting demonstrations of motion trajectories. However, in theory these two kinds of demonstrations could be used together in cases where the subtasks are difficult to learn without kinematic demonstrations. Given a task decomposition scheme, we then leverage human priors to label the correct sequence to achieve the subtasks. Regarding implementation, an abstract demonstration is represented by an ordered sequence of indexes, each corresponding to a subtask.

At the beginning of a learning episode, a final goal is sampled and given to the GRL agent. When abstract demonstrations are used, the agent is asked to learn the subtasks following the order given by the demonstration, instead of being directly given the final goal. The agent is given the next subtask if the current goal has been achieved. When given a new goal, the previous transitions within this episode are copied as a new trajectory and their desired goals are replaced with the new goal. New transitions associated with the new goals are then appended to the new trajectory, leaving the previous trajectory unchanged. This is to guarantee that the agent can finally learn to achieve all the goals without demonstrations. The episode ends when the number of maximum timesteps is reached.

We use a hyperparameter, $\eta \in [0,1]$, to control the proportion of training episodes that are demonstrated. We follow the training procedure used in \cite{Andrychowicz2017}, which is organised in epochs, cycles and episodes. Each epoch has a number of cycles, $I$, and each cycle has a number of episodes, $J$. The number of demonstrated episodes in a cycle is computed by $episode_{demo}=\eta \times J$.

\subsection{Adaptive exploration}

As mentioned above, the idea of adaptive exploration in multi-step task learning is to reduce unnecessary exploration for well-learnt steps and increase on unfamiliar steps. Coupled with abstract demonstrations, this leads to an adaptive curriculum that guarantees the GRL agent to proceed learning towards the final goal as quickly as possible. Reducing unnecessary exploration also leads to more stable performance. Since deterministic and stochastic agents explore differently, we provide different implementations for them as follows.

\textbf{\textit{Deterministic}} agents explore the environment using a base behavioural policy, $\pi_b(a|s,g)$, that contains some hyperparameters to control the exploratory behaviours. In this paper, we use the \textit{epsilon-Gaussian} (EGa) strategy as our base behavioural policy for the DDPG agent, and the \textit{epsilon-greedy} (EGr) strategy for the DQN agent. 

The EGr strategy takes a random action with a probability $\epsilon$ or takes an action that maximises the learnt Q-function. In practice, $\epsilon$ will decay to a lower bound during the course of training, commonly using the following equation:

\[
\epsilon = \epsilon_{end} + (\epsilon_{start} - \epsilon_{end}) \times e^{\frac{-n}{\beta}} \ \ \ \ 
\]

\noindent where, $\epsilon_{start}$ and $\epsilon_{end}$ are the upper and lower bounds, $n$ is the total elapsed environment timesteps and $\beta$ is the decay coefficient parameter. Intuitively, $\epsilon$ decays as the number of elapsed timesteps grows and its speed is controlled by $\beta$. To make it adaptive, we simply replace the exponential term by a performance metric. Specifically, we construct an $N$-dimensional vector to store the value of $\epsilon$ for each task step at the $m$-th epoch, $\boldsymbol{\epsilon}_m$. We then use an $N$-dimensional vector to record the test success rates for all task steps at the $m$-th epoch, denoted as $\boldsymbol{S}_m$, initialised to $\boldsymbol{0}$. We update $\boldsymbol{\epsilon}_m$ using the following equations:

\begin{equation}\label{eq:AE-EGr}
\boldsymbol{\epsilon}_{m+1} = \boldsymbol{\epsilon}_{end} + (\boldsymbol{\epsilon}_{start} - \boldsymbol{\epsilon}_{end}) \times (1-\boldsymbol{S}_m). \ \ \ \ 
\end{equation}

For continuous action cases, the EGa strategy samples a random action uniformly with a probability $\epsilon$ or takes an action generated by a learnt policy with noises from a fixed Gaussian distribution. That is,

\[
\pi_b(a|s,g)=
\begin{cases}
	a \sim \mathcal{U}(|\mathcal{A}|, -|\mathcal{A}|),  & \delta \leq \epsilon \\
	a \sim \mathcal{N}(\pi(a|s,g), \sigma),               & \delta > \epsilon
\end{cases}
\]
where $\delta\sim\mathcal{U}(0,1)$ and $\pi(a|s,g)$ is a learnt, deterministic, goal-conditioned policy. By varying the hyperparameters, $\epsilon$ and $\sigma$, based on the performance of each task step at different training epochs, we have an adaptive exploration strategy. Again, we use $N$-dimensional vectors to represent these two hyperparameters for the $N$ task steps at the $m$-th epoch, $\boldsymbol{\epsilon}_m$ and $\boldsymbol{\sigma}_m$. At the beginning of a training process, these two vectors are initialised to their initial values, $\boldsymbol{\epsilon}_0$ and $\boldsymbol{\sigma}_0$ ($\boldsymbol{\epsilon}_0=0.2$ and $\boldsymbol{\sigma}_0=0.05$ in all our experiments). We use another vector to record the test success rates and update $\boldsymbol{\epsilon}_m$ and $\boldsymbol{\sigma}_m$ by:

\begin{equation}\label{eq:AE-update-rule}
\boldsymbol{\epsilon}_{m+1} = \boldsymbol{\epsilon}_0 \times (1-\boldsymbol{S}_m),\ \ \ \ 
\boldsymbol{\sigma}_{m+1} = \boldsymbol{\sigma}_0 \times (1-\boldsymbol{S}_m).
\end{equation}

\textbf{\textit{Stochastic}} agents explore the environment by sampling from a learnt stochastic policy. As an example, the SAC agent uses a Gaussian policy whose mean and deviation are produced by a neural network. We propose to use the success rate as a scaling factor for the deviation for different task steps. Similar to Eq.~\ref{eq:AE-update-rule}:

\begin{equation}\label{eq:AE-update-rule-stochastic}
\boldsymbol{\sigma}_{m+1} = \boldsymbol{\sigma}_0 \times (1-\boldsymbol{S}_m),
\end{equation}

where $\boldsymbol{\sigma}_m$ is the deviations for the learnt stochastic policy for different task steps, and $\boldsymbol{\sigma}_0$ can be pre-defined constants or learnt by a neural network. 

\textbf{\textit{Performance metrics.}} To obtain the test success rates, we perform $K$ testing episodes for each task step after each training epoch and calculate the average success rate for each step. This test is run with abstract demonstrations to reflect the performance of achieving a step starting from the previous one. However, computing Eqs.~\ref{eq:AE-EGr}, \ref{eq:AE-update-rule} and \ref{eq:AE-update-rule-stochastic} with the vanilla success rate would result in bumping changes of the exploration hyperparameters. As the success rate is calculated only from the $K$ testing episodes, it may not reflect the true performance of the current policy, because $K$ is usually not large for the sake of reducing computations. Thus, we use the polyak-average \cite{scieur2020universal} of the test success rate vector, $\boldsymbol{S}_m^-$ to calculate Eqs.~\ref{eq:AE-EGr}, \ref{eq:AE-update-rule} and \ref{eq:AE-update-rule-stochastic} instead of the vanilla success rates. The running average is calculated with a parameter $\tau_{\boldsymbol{S}} \in [0, 1]$ as follows:

\begin{equation}\label{eq:sr-polyak}
\boldsymbol{S}_{m}^- = (1-\tau_{\boldsymbol{S}})\times \boldsymbol{S}_{m-1}^- + \tau_{\boldsymbol{S}} \times \boldsymbol{S}_{m}
\end{equation}

To demonstrate the proposed $\mathbf{A^2}$ method, we incorporate it with the DQN, DDPG and the SAC agents. For the DQN and DDPG agents, we use the EGr and EGa as the base methods for adaptive exploration and use Eqs.~\ref{eq:AE-EGr} and \ref{eq:AE-update-rule} to update their parameters. For the SAC agent, we use Eq.~\ref{eq:AE-update-rule-stochastic} to scale the policy deviation learnt by a neural network.

\section{Results}
\label{sec:result}

\begin{figure}[h]
    \centering
	\subfloat[\label{fig:gridworld}GridDoorKey]{
	\includegraphics[height=0.11\textwidth]{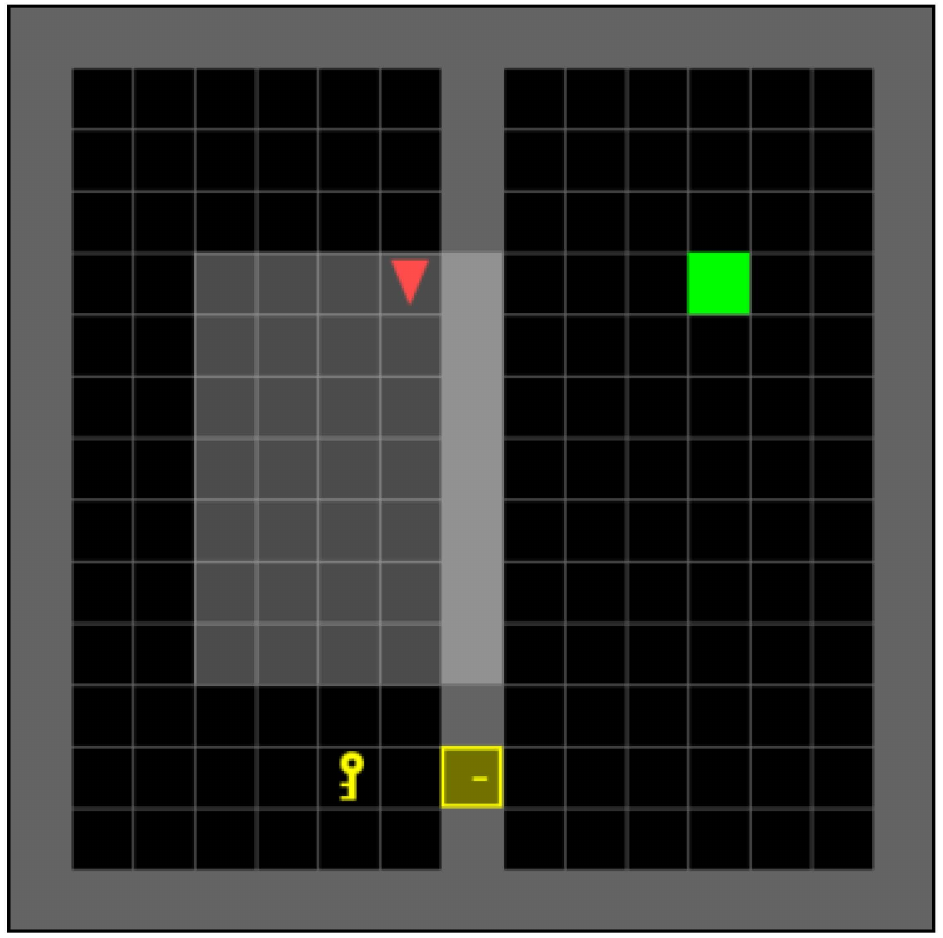}}
    \hfill
    \subfloat[\label{fig:chestpush}ChestPush]{
    \includegraphics[height=0.11\textwidth]{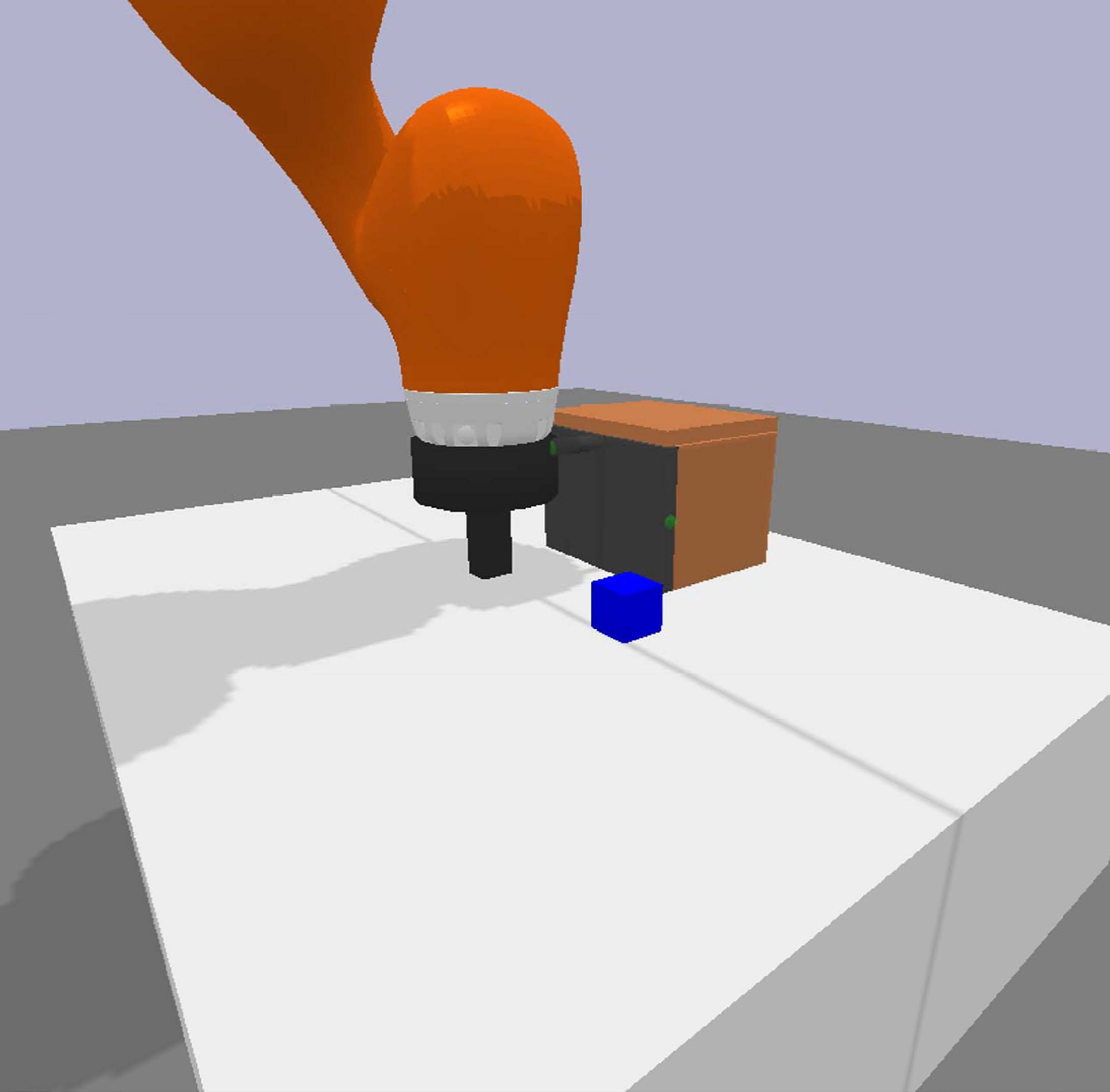}}
	\hfill
	\subfloat[\label{fig:chestpick}ChestPick]{
	\includegraphics[height=0.11\textwidth]{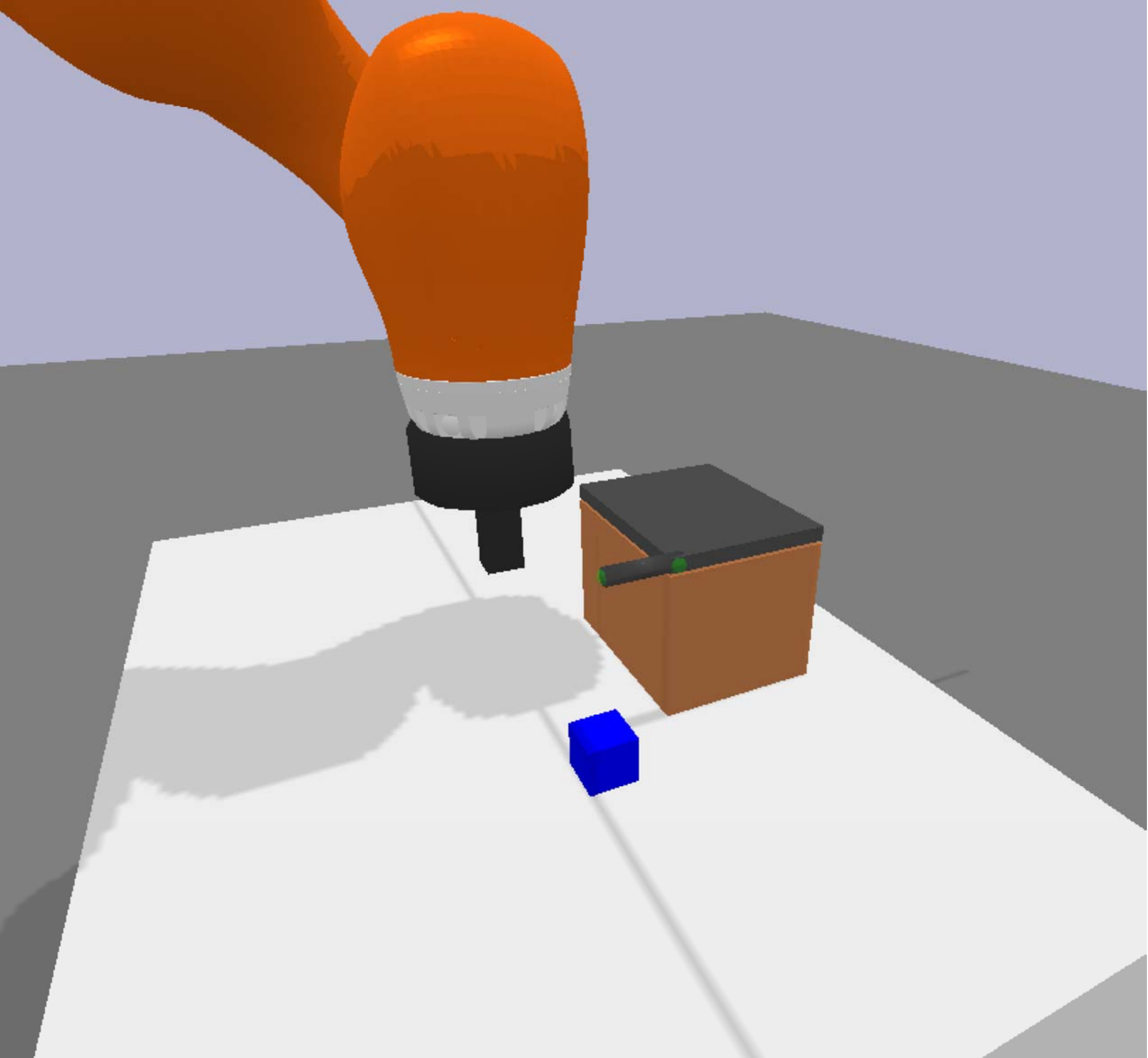}}
	\hfill
	\subfloat[\label{fig:blockstack}BlockStack]{
	\includegraphics[height=0.11\textwidth]{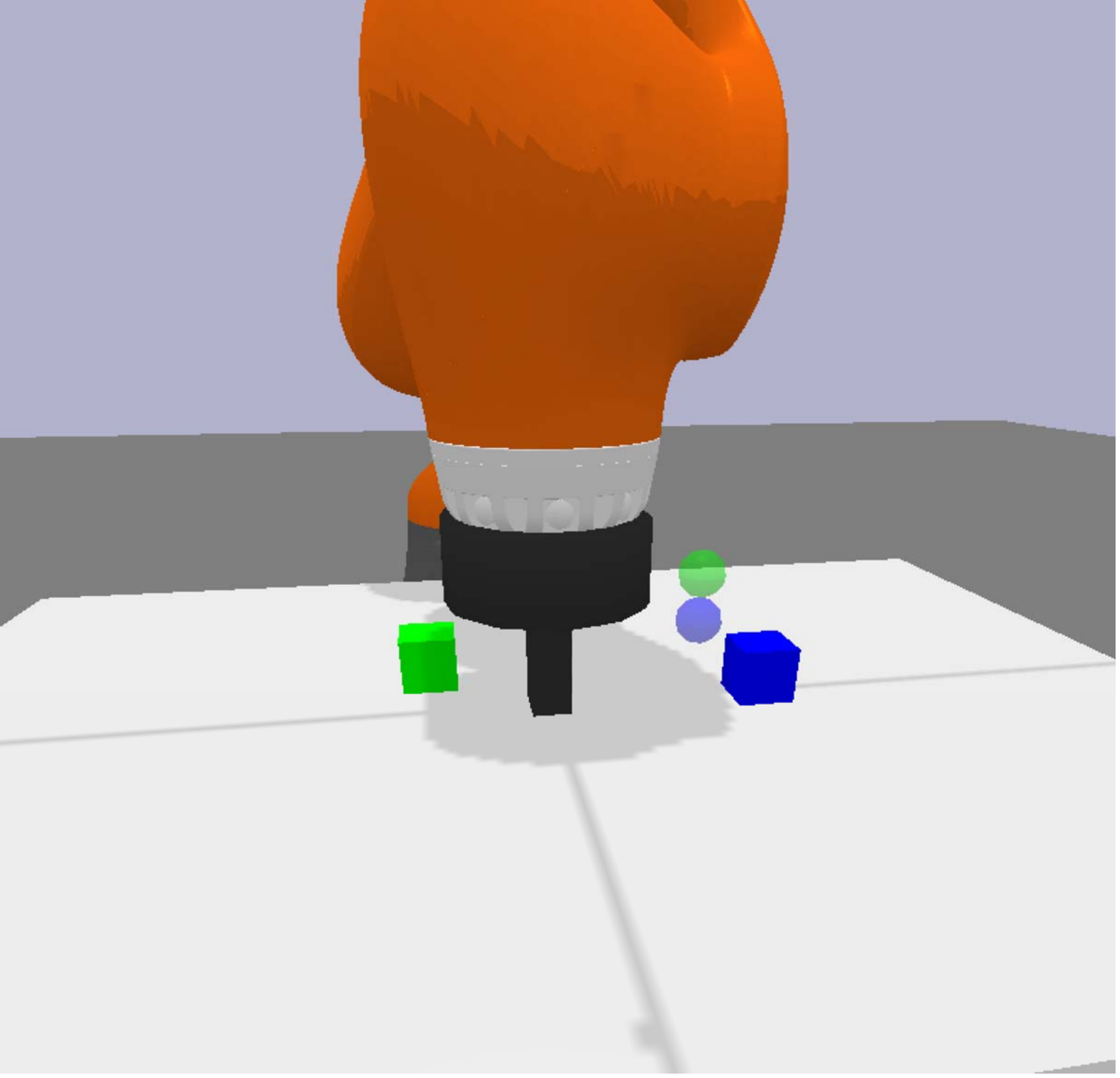}}
	\caption{Tasks. (a) The agent (red) needs to reach the goal (green). (b-c) The robot needs to push or pick-and-drop the block into the chest. (d) The robot needs to stack the blocks. \label{fig:tasks}}
\end{figure}

To investigate the effect of the proposed $\mathbf{A^2}$ method, we conduct a series of simulation experiments using the Mini-Grid \cite{gym_minigrid} and PMG environments \cite{yang2021open}. All performances displayed were the success rates of achieving the final goal without demonstrations, averaged over five random seeds. Specifically, we experimented on six multi-step, sparse reward tasks, including GridDoorKey (3 sizes), ChestPush, ChestPickAndPlace and BlockStack (see Fig.~\ref{fig:tasks} for a visualisation). Codes available at \url{https://github.com/IanYangChina/A-2-paper-code}.

\subsection{Task and implementation details}\label{subsec:task-detail}

\textbf{GridDoorKey task}: The state representation consists of the $x$ and $y$ coordinates of the agent, the key and the door, the heading direction of the agent, and two binary variables indicating whether the agent is carrying the key and whether the door is opened. The goal representation is the $x$ and $y$ coordinates of the target. The reward function gives a value of $0$ when the goal is reached and $-1$ otherwise.

\textbf{Manipulation tasks:} We use the ChestPush and ChestPick tasks with one block and the BlockStack task with two blocks from the Pybullet Multigoal (PMG) environments \cite{yang2021open}. The state representation and reward function remain the same as the original paper. The original goal representation only consists of the target coordinates of the blocks in the world frame. We add the gripper location and its finger width to the goal representation for specifying the task steps regarding grasping and placing.

At the beginning of each task, the agent is given a desired goal which, for example, specifies the desired location of a block. If abstract demonstrations are used in an episode, the agent will instead be given a subgoal according to the step at hand. The decomposition schemes for the tasks are as follows:

\begin{itemize}
    \item Gridworld: three subgoals, including 1) reach the key, 2) reach the door and 3) reach the target location;
    \item ChestPush: three subgoals, including 1) open the chest, 2) reach the block and 3) push the block into the chest;
    \item ChestPick: four subgoals, including 1) open the chest, 2) grasp the block, 3) move to the top of the chest and 4) drop the block into the chest;
    \item BlockStack: four subgoals, including 1) grasp the base block, 2) move the base block to the target location, 3) grasp the second block and 4) stack the second block.
\end{itemize}

\textbf{Network architecture:} The DQN network has three MLPs of sizes 64, 128 and 64. The actor and critic networks for the DDPG and SAC have three MLPs of sizes 256. All layers use ReLU activation, except for the output. All actors use hyperbolic tangent to activate the final layer and all critics (include DQN) have no activation on the output.

\subsection{Ablations}\label{subsec:ablation}

This subsection examines the effect of different parameter values of the $\mathbf{A^2}$ method, namely, the percentage of demonstrated episodes $\eta$ and the adaptive exploration update rate $\tau_{\boldsymbol{S}}$. We perform ablations on the GridDoorKey25x25 tasks with the DQN agent, ChestPush task with the DDPG agent and BlockStack task with the SAC agent.

\begin{figure}[t]
	%\vspace{-5pt}
    \centering
	\subfloat[\label{subfig:ta-eta-grid}Gridworld25-DQN]{
	\includegraphics[height=0.125\textwidth]{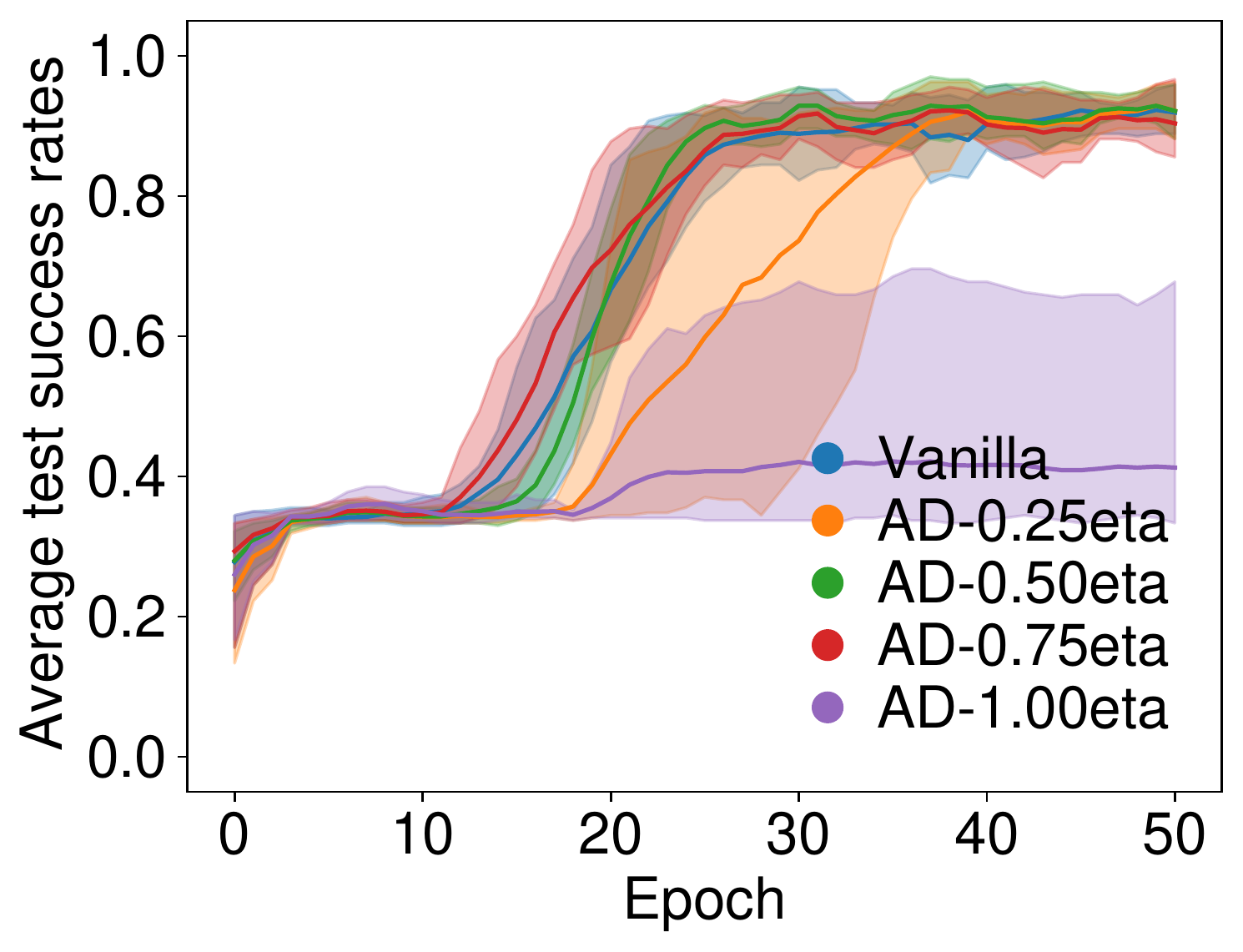}}
	\hfil
	\subfloat[\label{subfig:ta-eta-ddpg-cp}ChestPush-DDPG]{
	\includegraphics[height=0.125\textwidth]{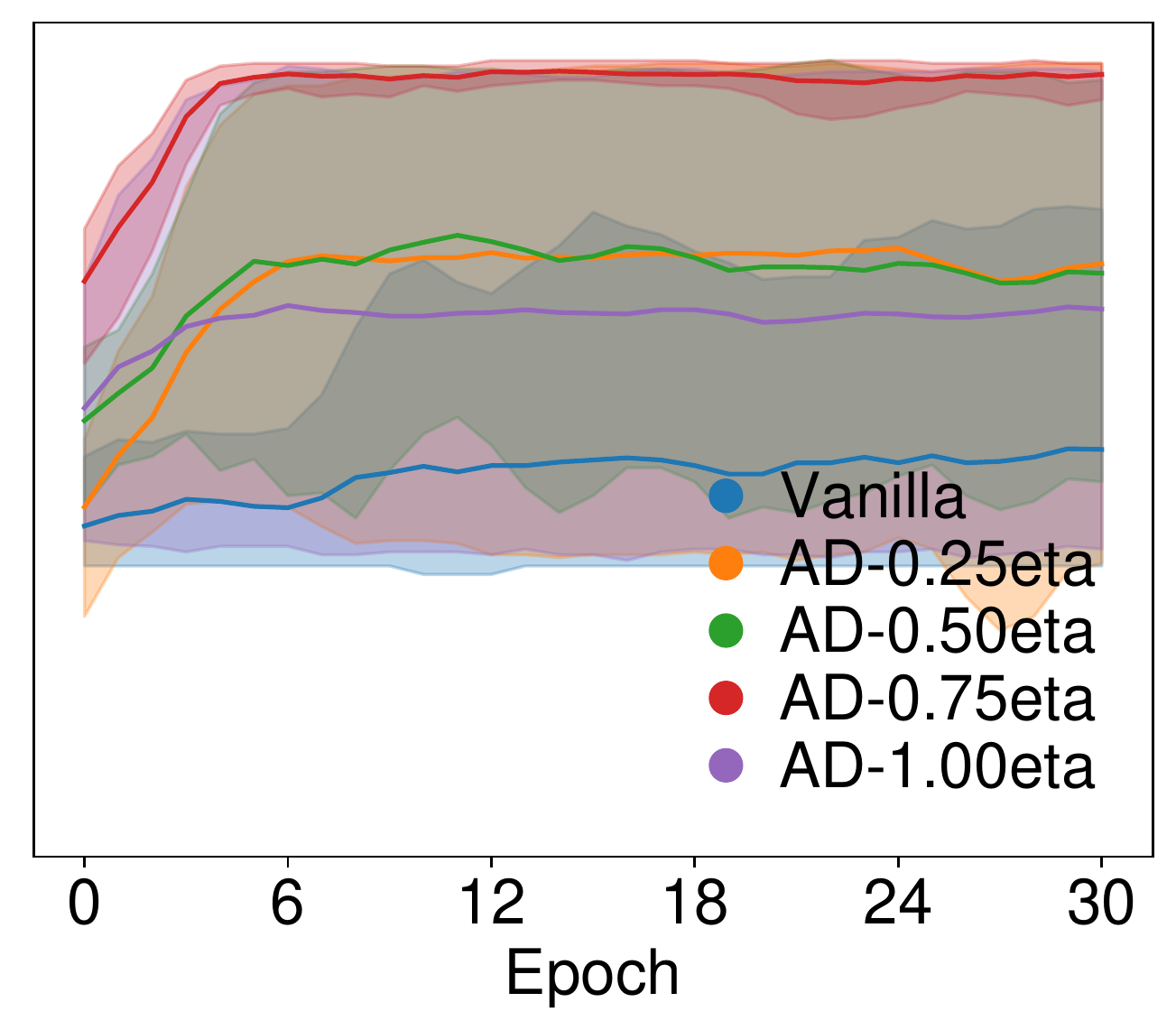}}
	\hfil
	\subfloat[\label{subfig:ta-eta-sac-cp}BlockStack-SAC]{
	\includegraphics[height=0.125\textwidth]{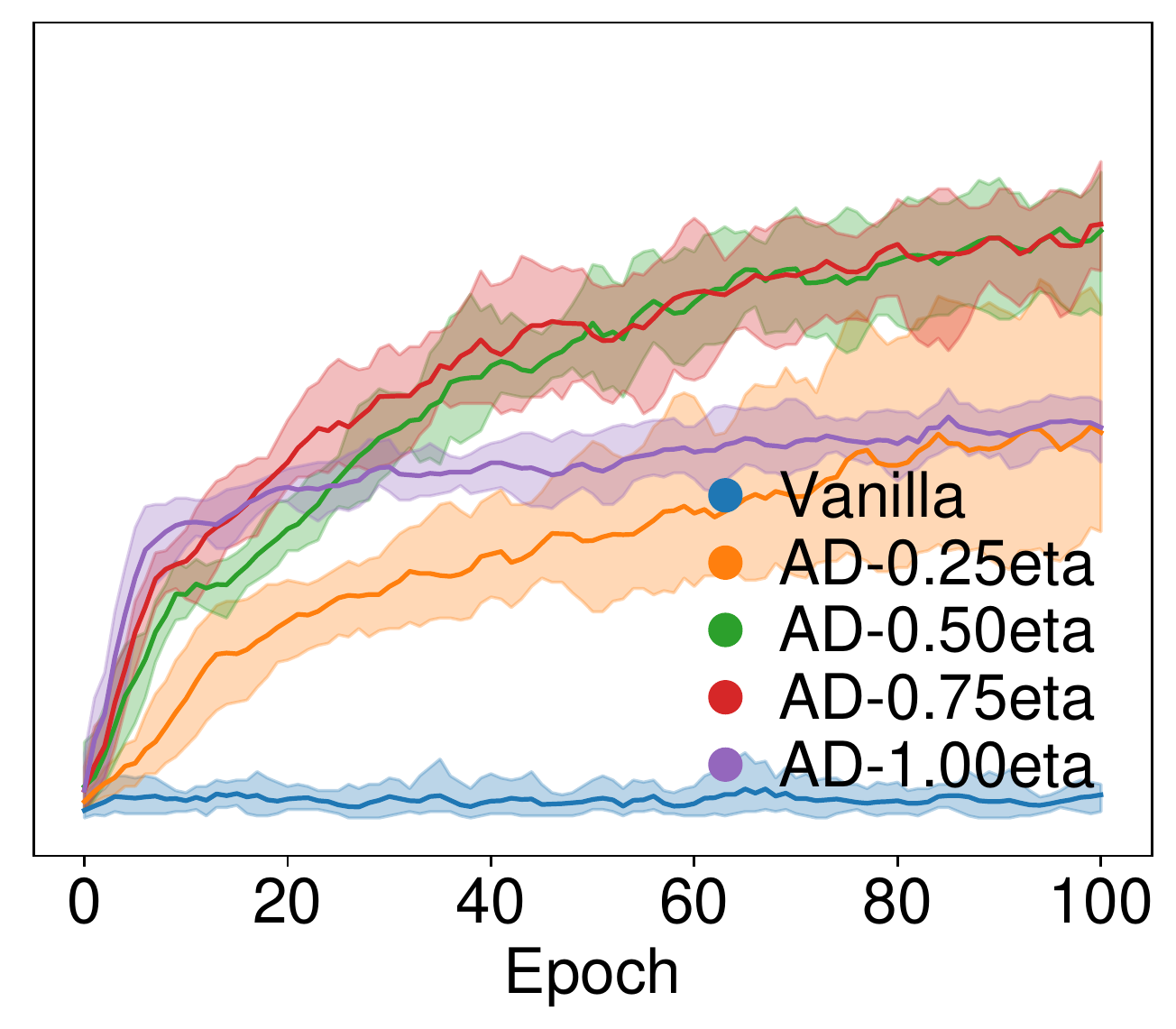}}\\
	
    \centering
	\subfloat[\label{subfig:ta2-tau-grid}Gridworld25-DQN]{
	\includegraphics[height=0.125\textwidth]{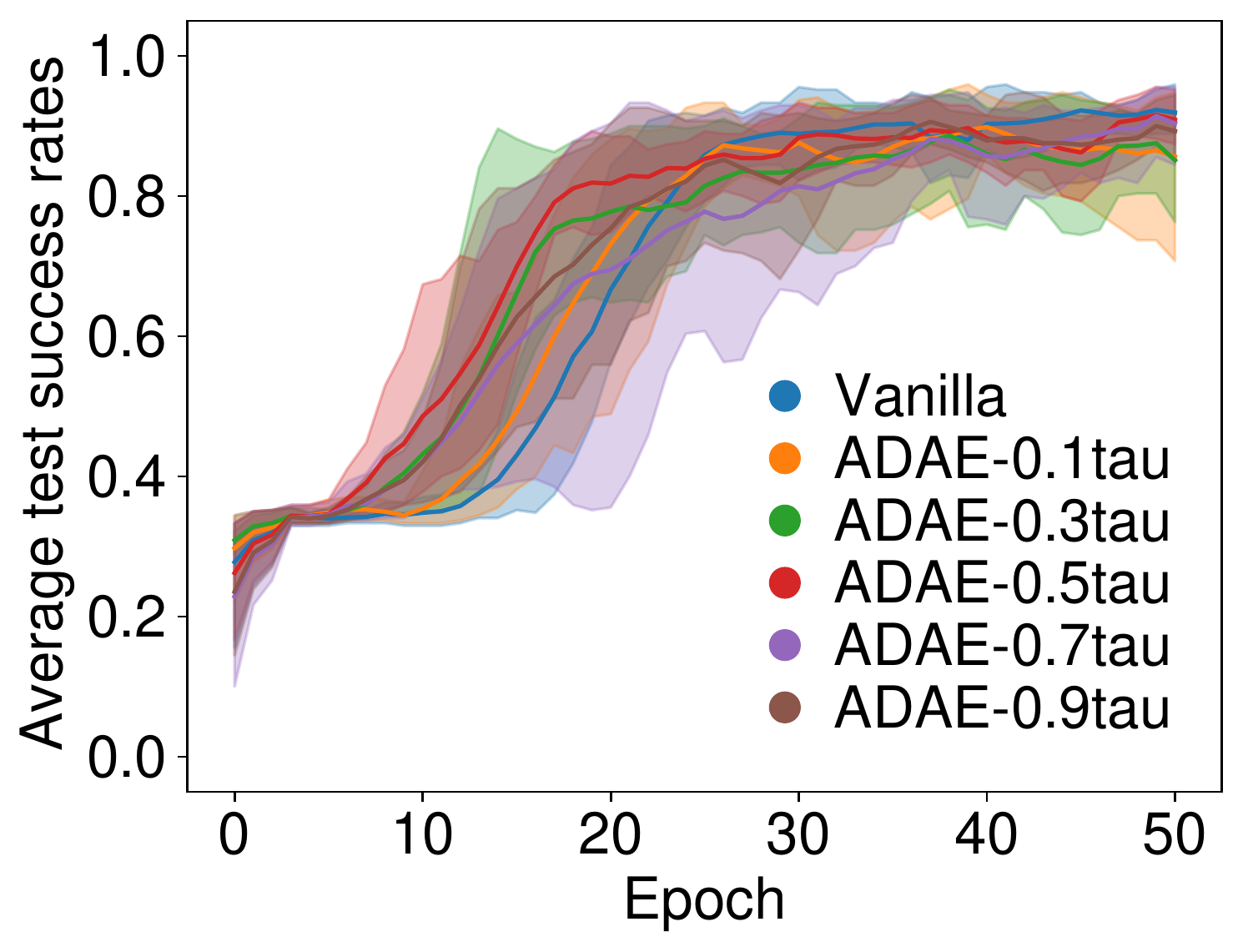}}
	\hfil
	\subfloat[\label{subfig:ta2-tau-ddpg-cp}ChestPush-DDPG]{
	\includegraphics[height=0.125\textwidth]{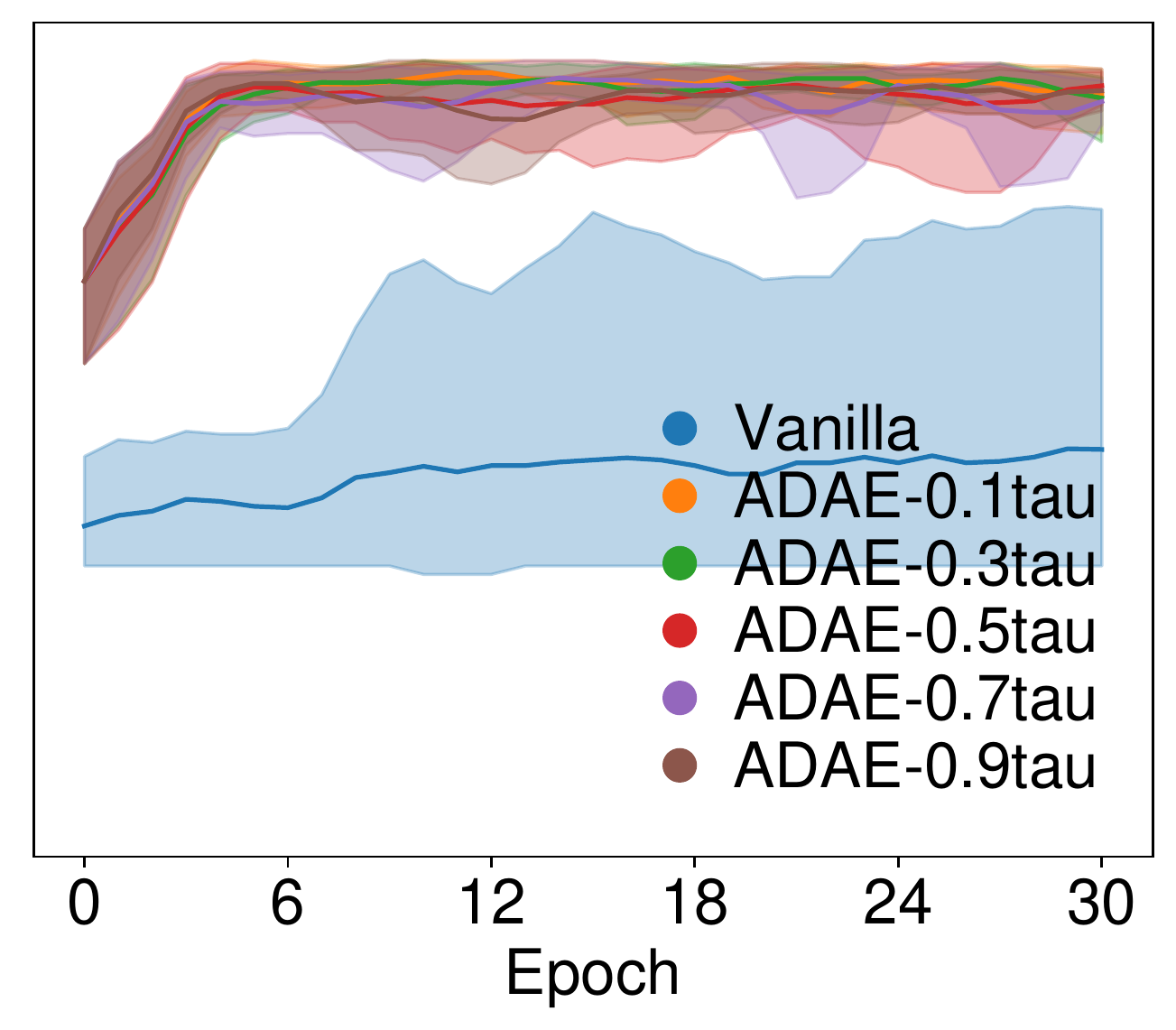}}
	\hfil
	\subfloat[\label{subfig:ta2-tau-sac-cp}BlockStack-SAC]{
	\includegraphics[height=0.125\textwidth]{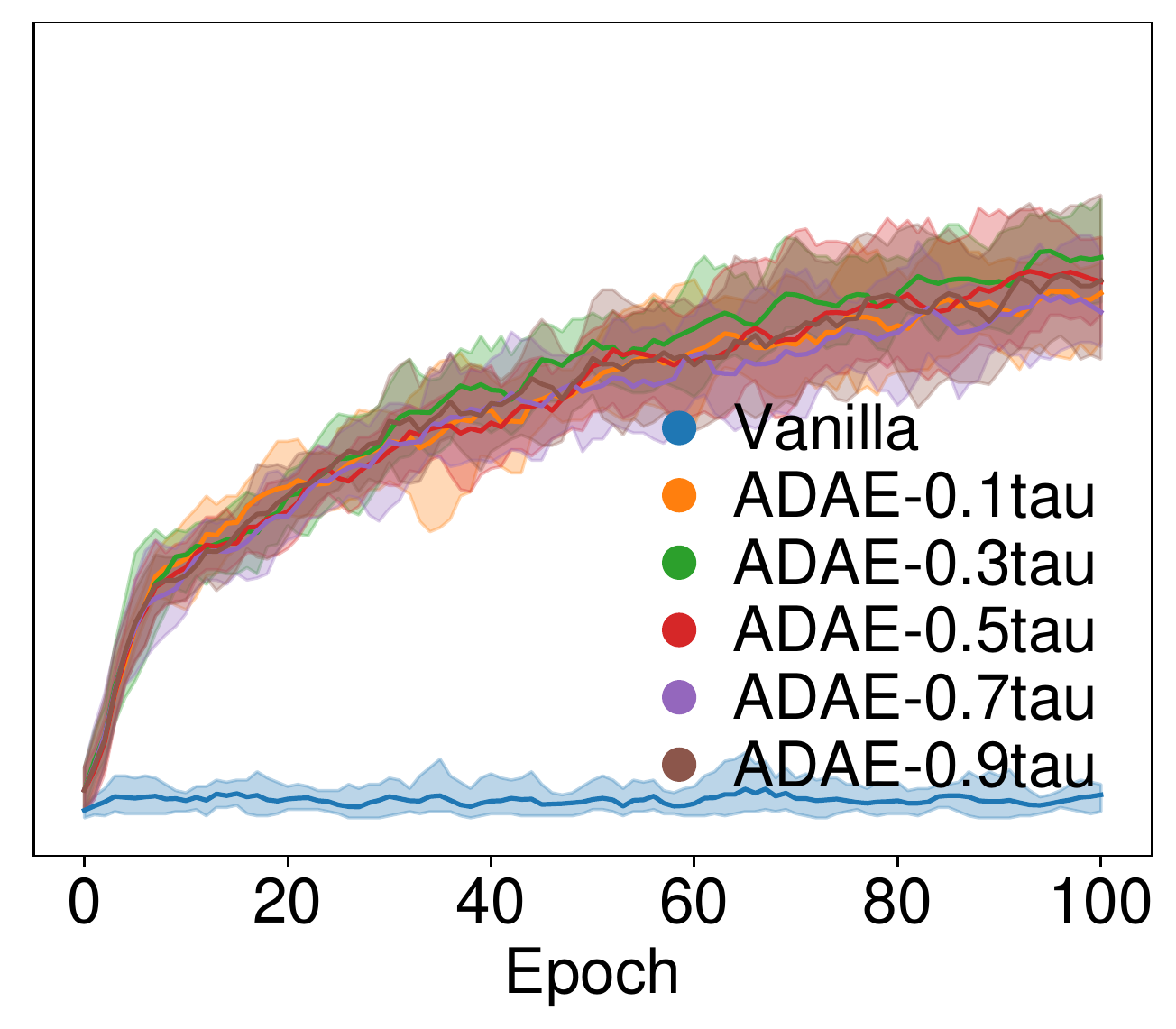}}
	\caption{Test success rates of achieving the final goal with different proportions of demonstrated episodes $\eta$ (a-c) and adaptive exploration update rate $\tau_{\boldsymbol{S}}$ (d-f). \textit{AD}: abstract demonstrations; \textit{ADAE}: abstract demonstrations and adaptive exploration.
	\label{fig:ablation}}
	\vspace{-15pt}
\end{figure}

From Figs.~\ref{subfig:ta-eta-grid} to \ref{subfig:ta-eta-sac-cp}, we see that adding demonstrations improves convergence speeds and performances in general. Adding demonstrations to $50\%$ or $75\%$ of training episodes has the highest performance gains across all agents and tasks (green and red lines). Interestingly, providing demonstrations to all the training episodes tends to hurt the performances (purple lines). This could be due to a lack of negative learning signals when trained in a fully-demonstrated way, given that the value function normally requires negative examples to distinguish good transitions from bad ones for more accurate value prediction.

The ablations of the adaptive exploration update rate were conducted with a fixed percentage of demonstrated episodes ($\eta=0.75$). From Figs.~\ref{subfig:ta2-tau-grid} to \ref{subfig:ta2-tau-sac-cp}, we see that, varying the value of $\tau_{\boldsymbol{S}}$ tends to have no obvious effects. However, a small value may slow down learning as shown by the gridworld experiment (the orange line in Fig.~\ref{subfig:ta2-tau-grid}).

\subsection{General performance}\label{subsec:gp}

This subsection examines the general improvements gained by the proposed $\mathbf{A^2}$ method. According to the ablation studies (subsection~\ref{subsec:ablation}), we add abstract demonstrations into $75\%$ of the training episodes and use $0.3$ for the adaptive exploration parameter, $\tau_{\boldsymbol{S}}$, for improvement evaluation.

\begin{figure}[t]
	%\vspace{-15pt}
    \centering
	\subfloat[\label{subfig:15gw}GridDoorKey15]{
	\includegraphics[height=0.126\textwidth]{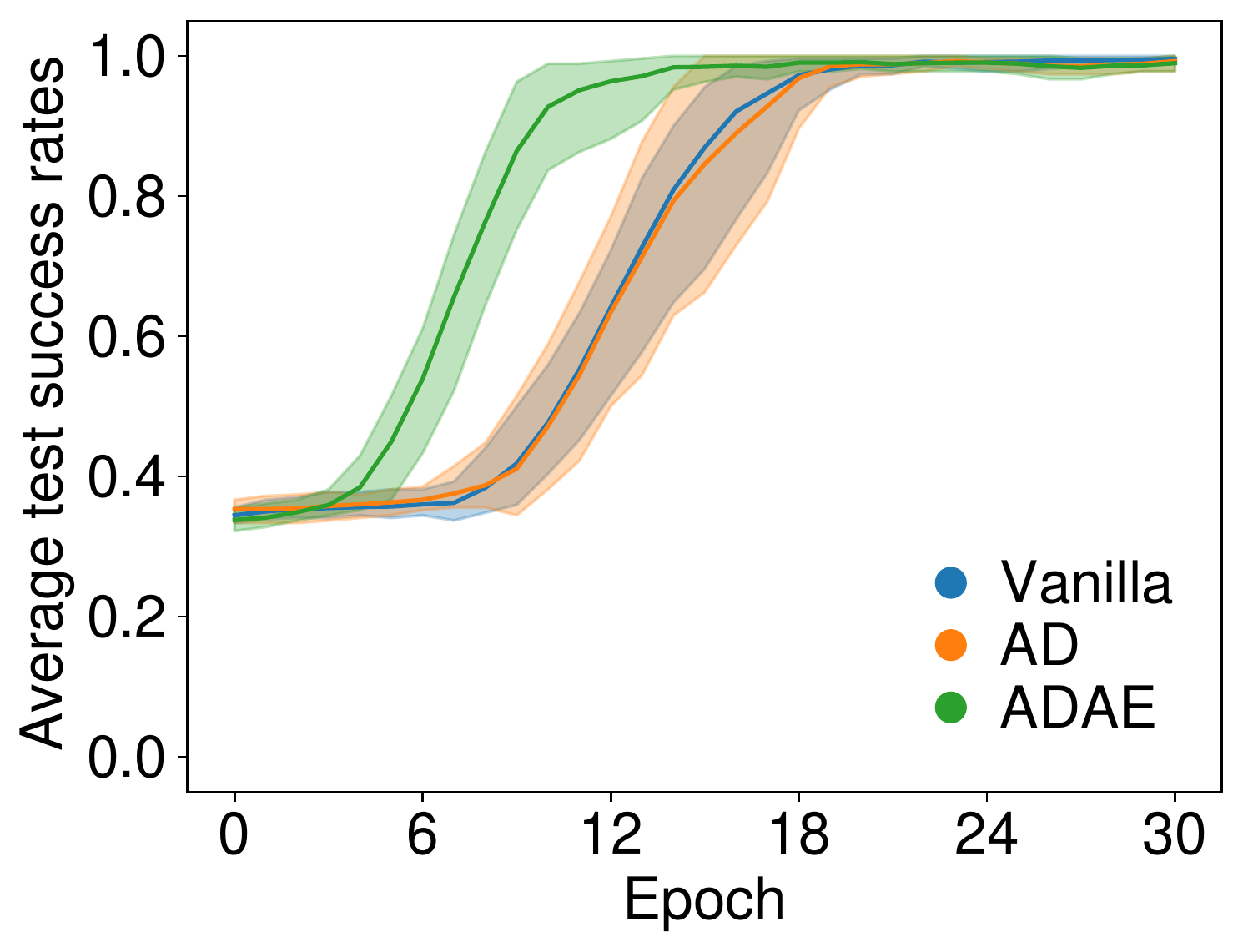}}
    \hfil
    \subfloat[\label{subfig:25gw}GridDoorKey25]{
    \includegraphics[height=0.126\textwidth]{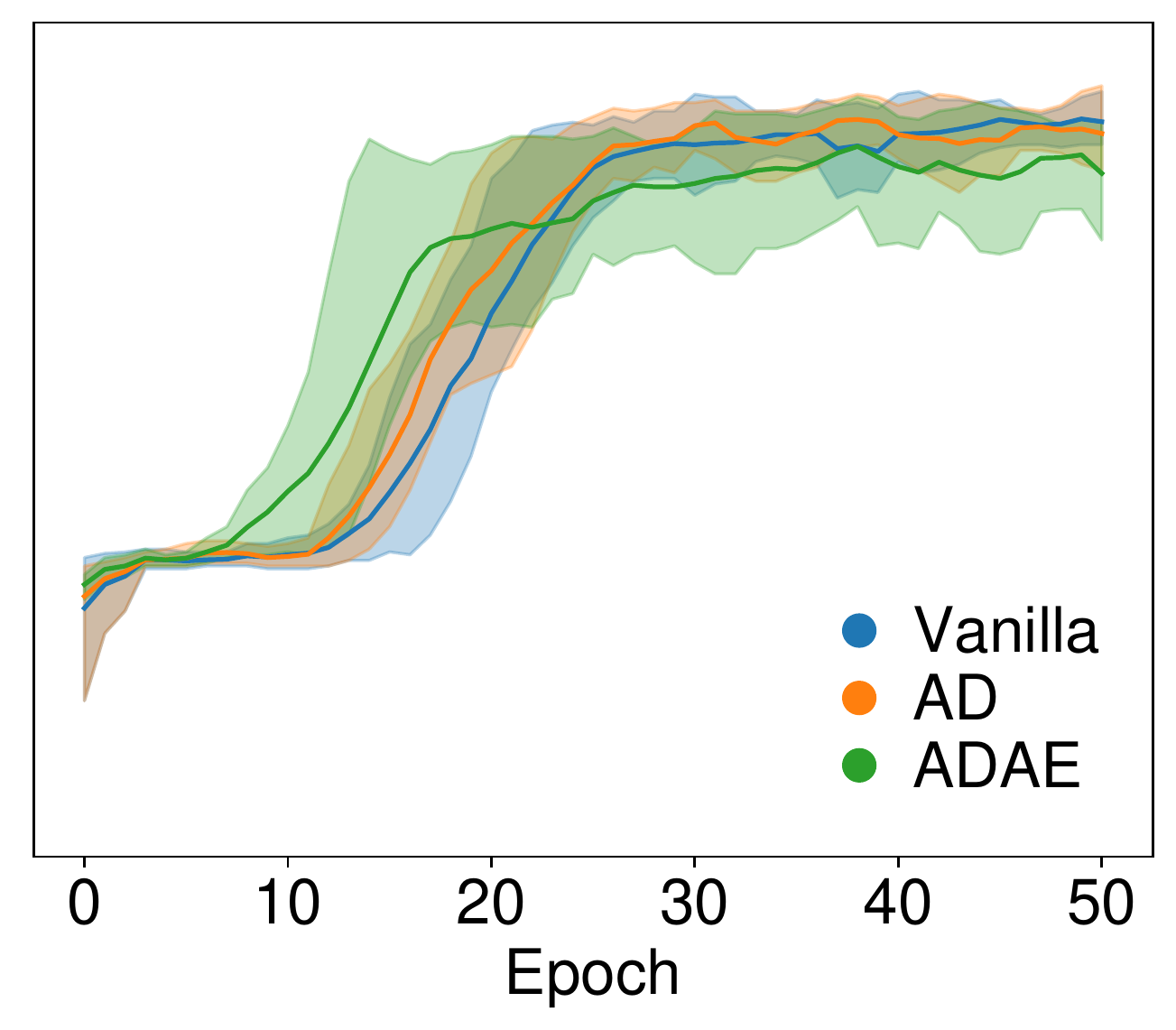}}
    \hfil
    \subfloat[\label{subfig:35gw}GridDoorKey35]{
    \includegraphics[height=0.126\textwidth]{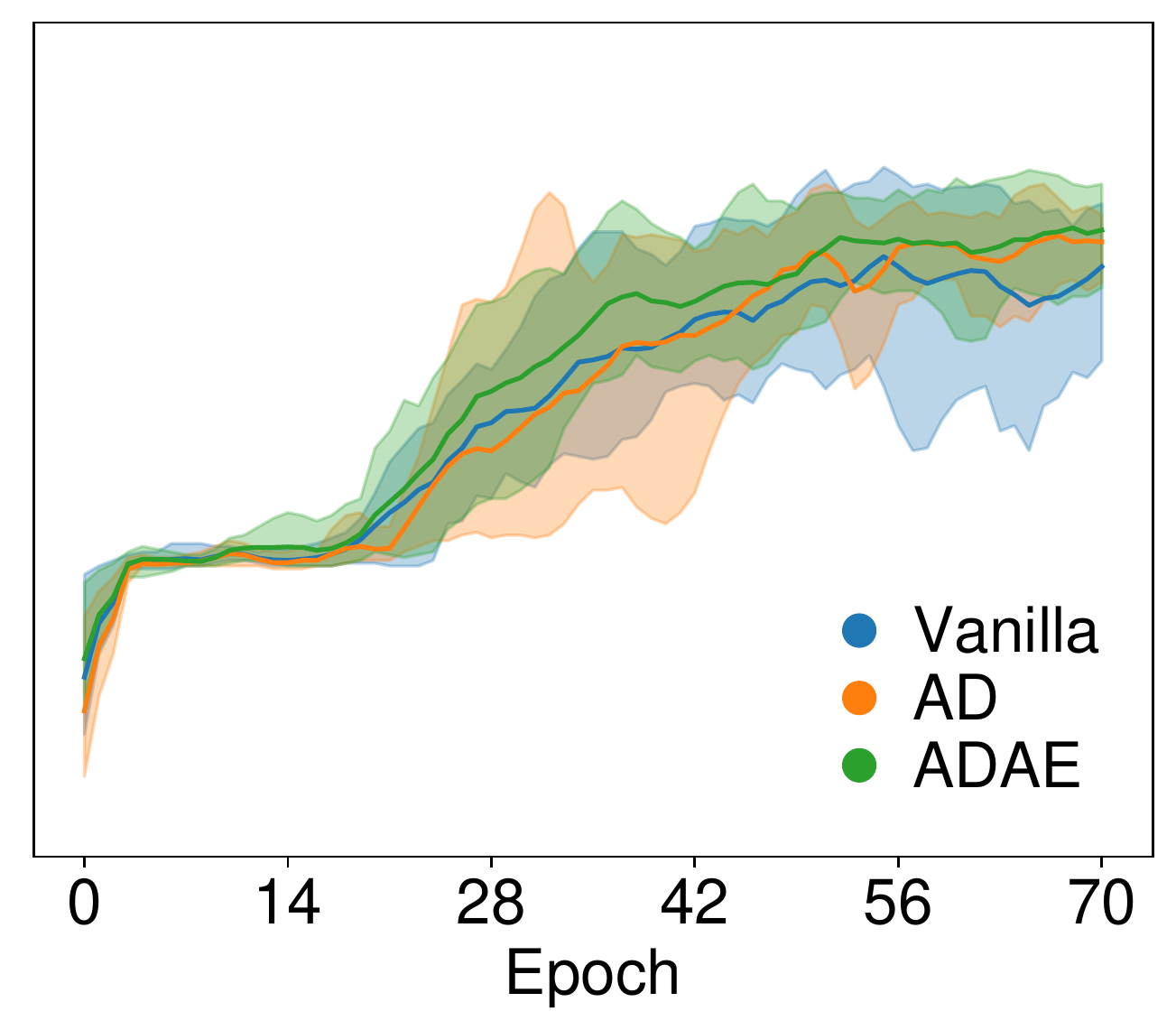}}\\
	
    \centering
	\subfloat[\label{subfig:ddpg-cp}ChestPush]{
	\includegraphics[height=0.126\textwidth]{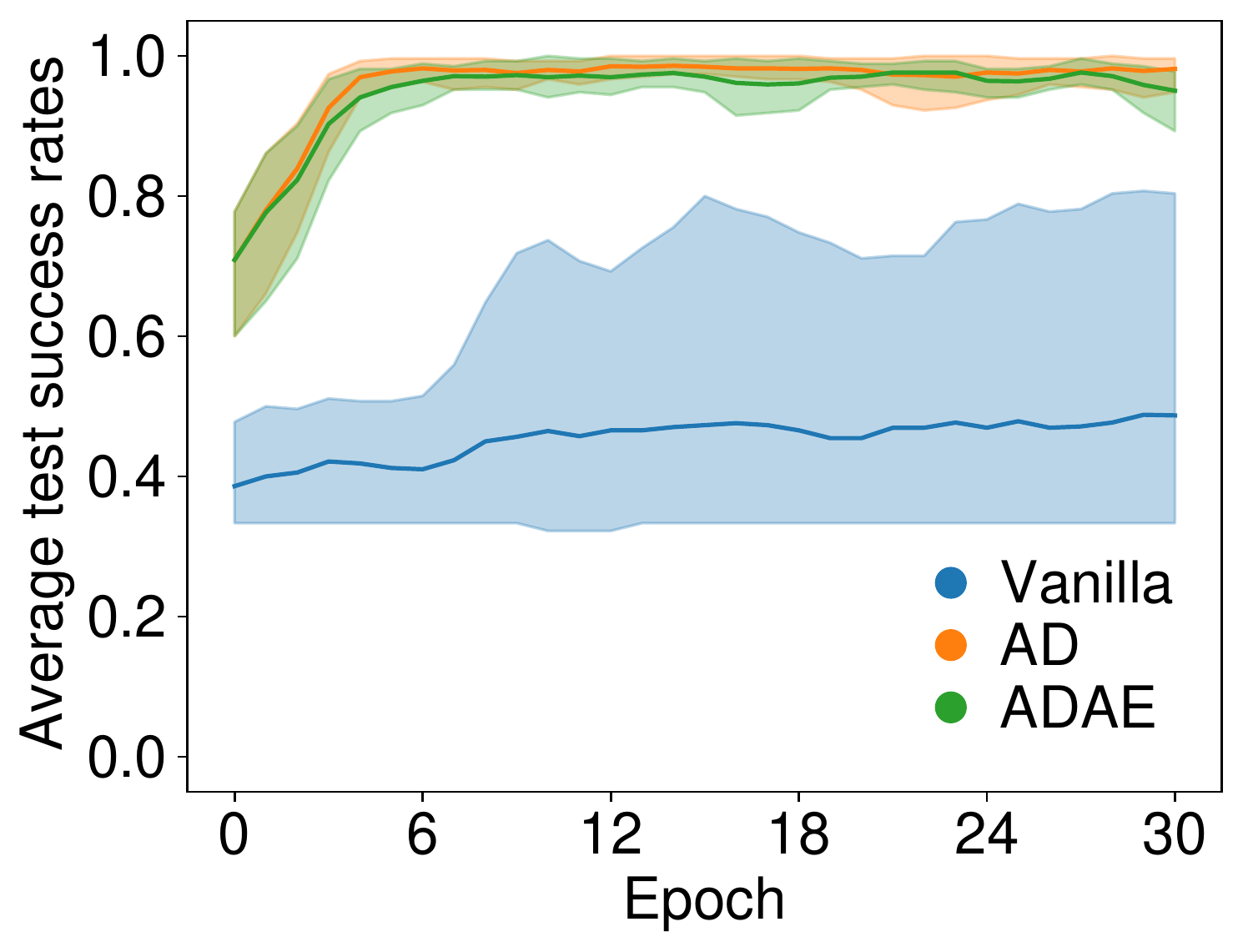}}
    \hfil
    \subfloat[\label{subfig:ddpg-ccp}ChestPickAndPlace]{
    \includegraphics[height=0.126\textwidth]{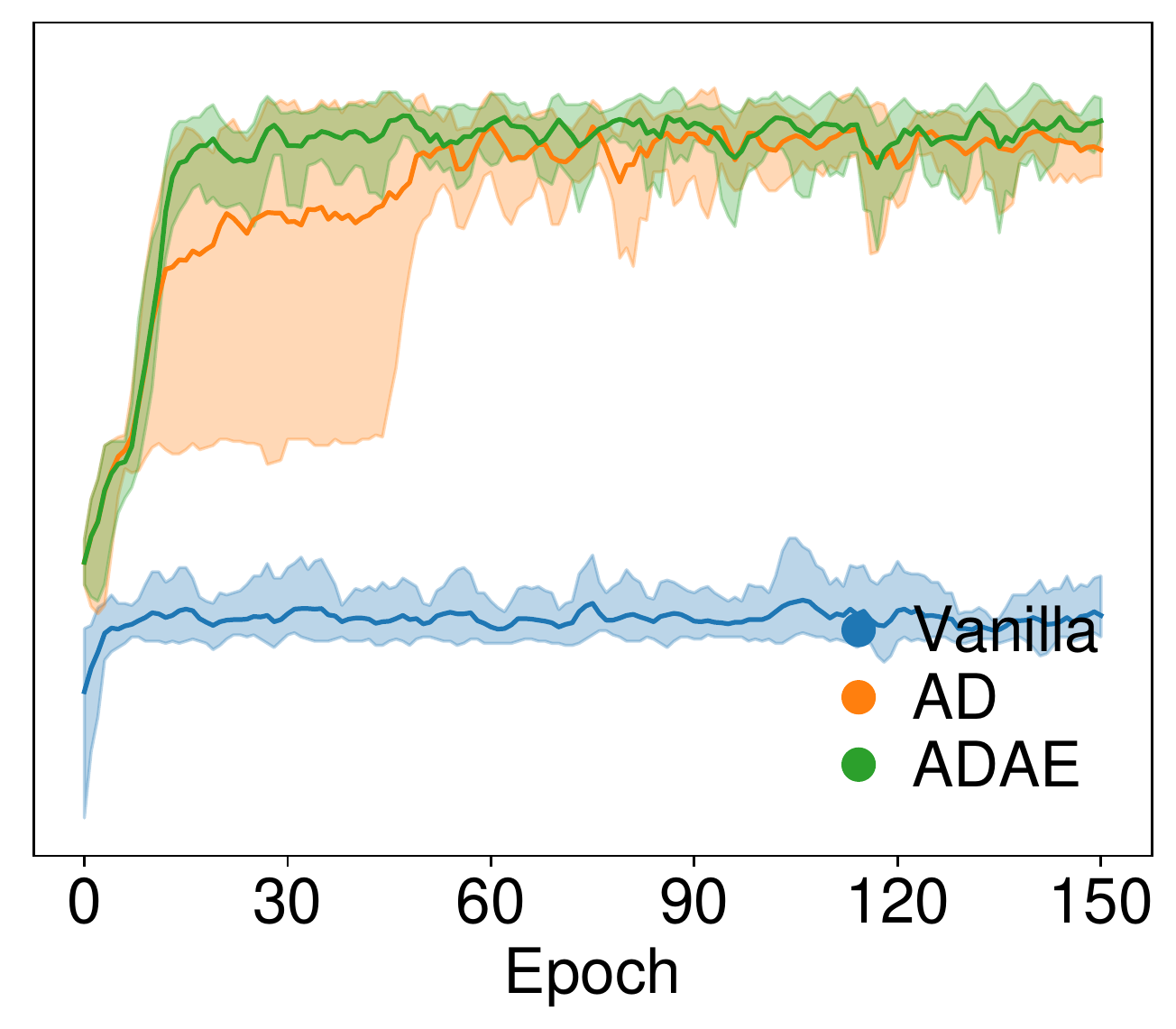}}
	\hfil
	\subfloat[\label{subfig:ddpg-bs}BlockStack]{
	\includegraphics[height=0.126\textwidth]{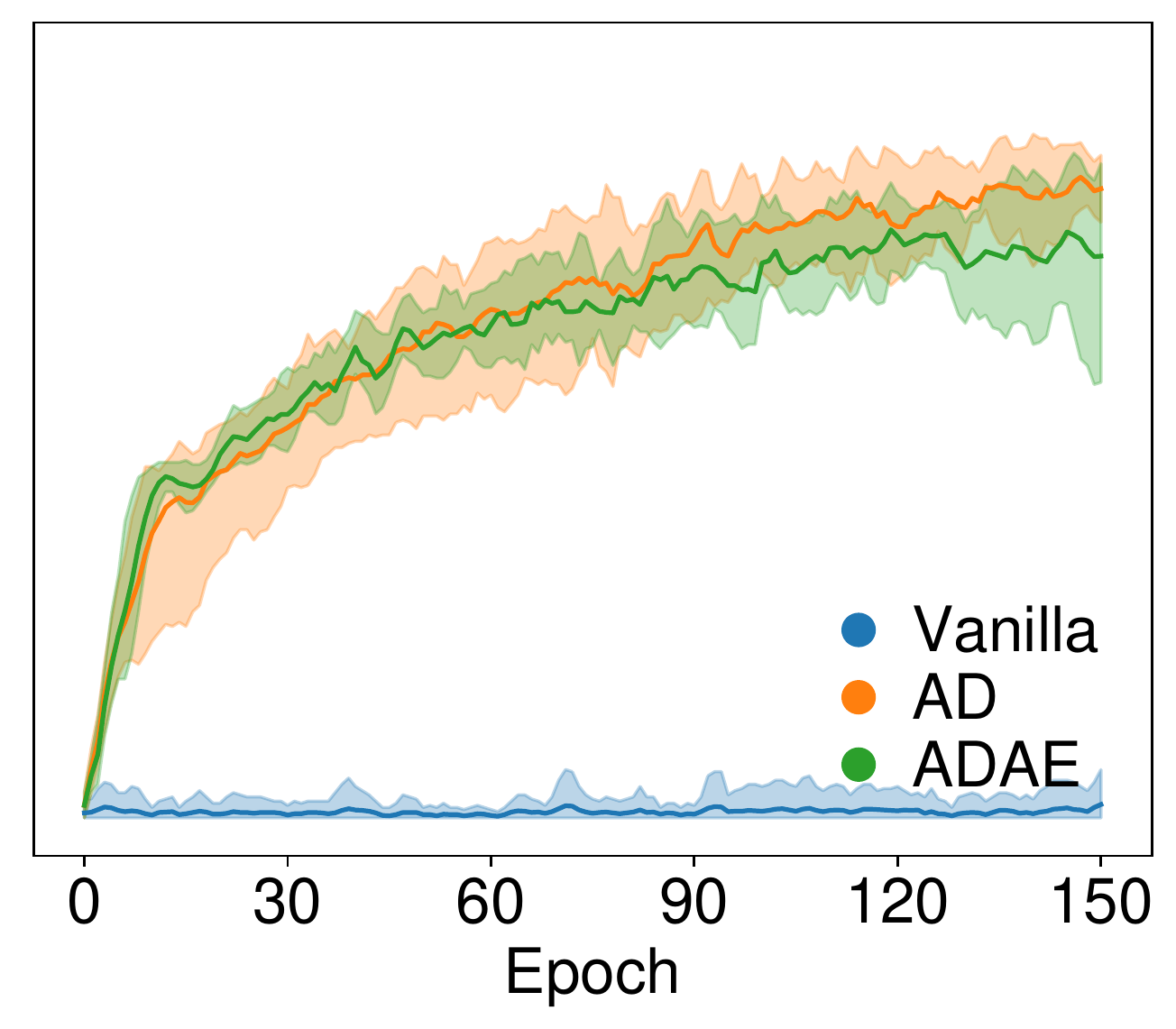}}\\
	
    \centering
	\subfloat[\label{subfig:sac-cp}ChestPush]{
	\includegraphics[height=0.126\textwidth]{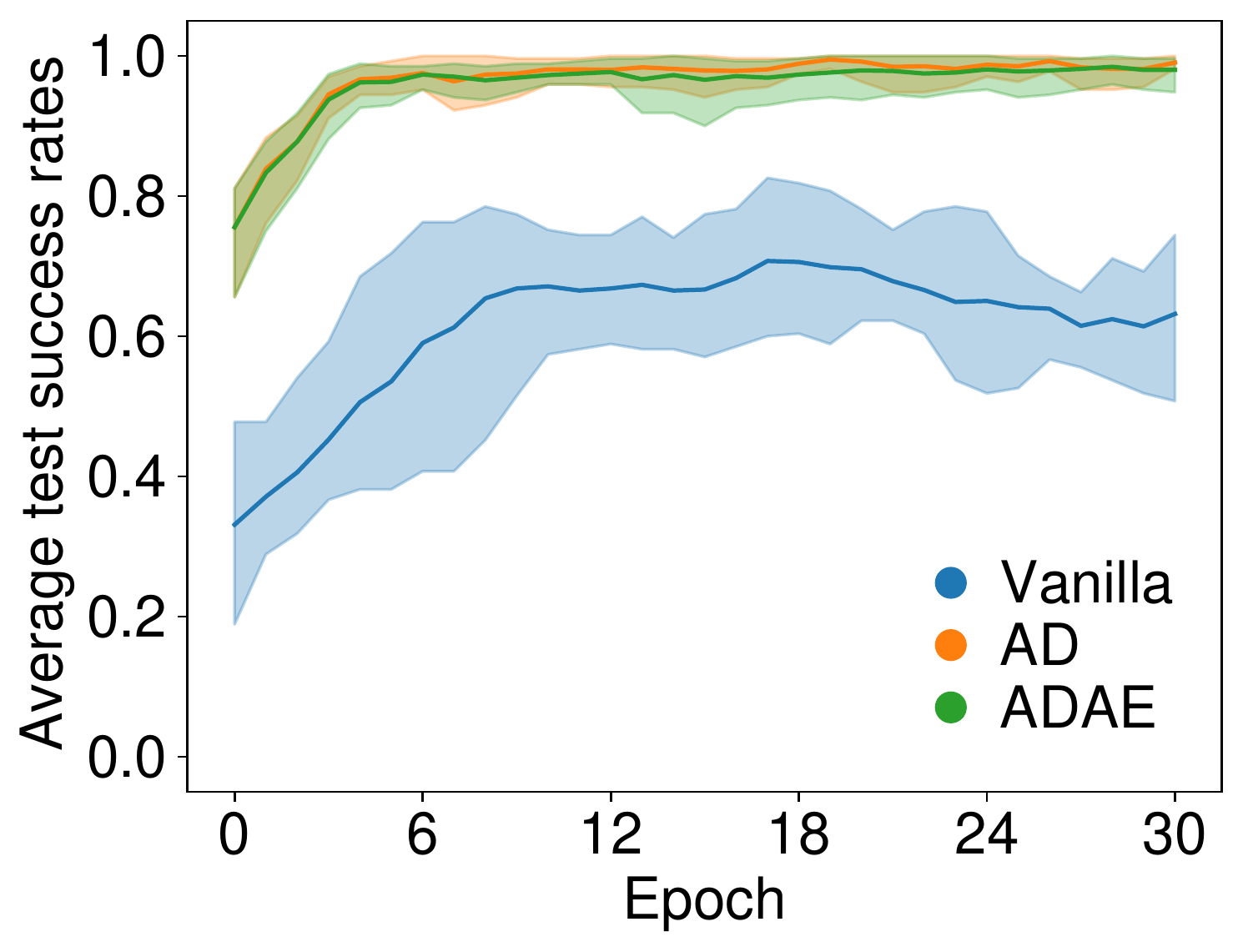}}
    \hfil
    \subfloat[\label{subfig:sac-ccp}ChestPickAndPlace]{
    \includegraphics[height=0.126\textwidth]{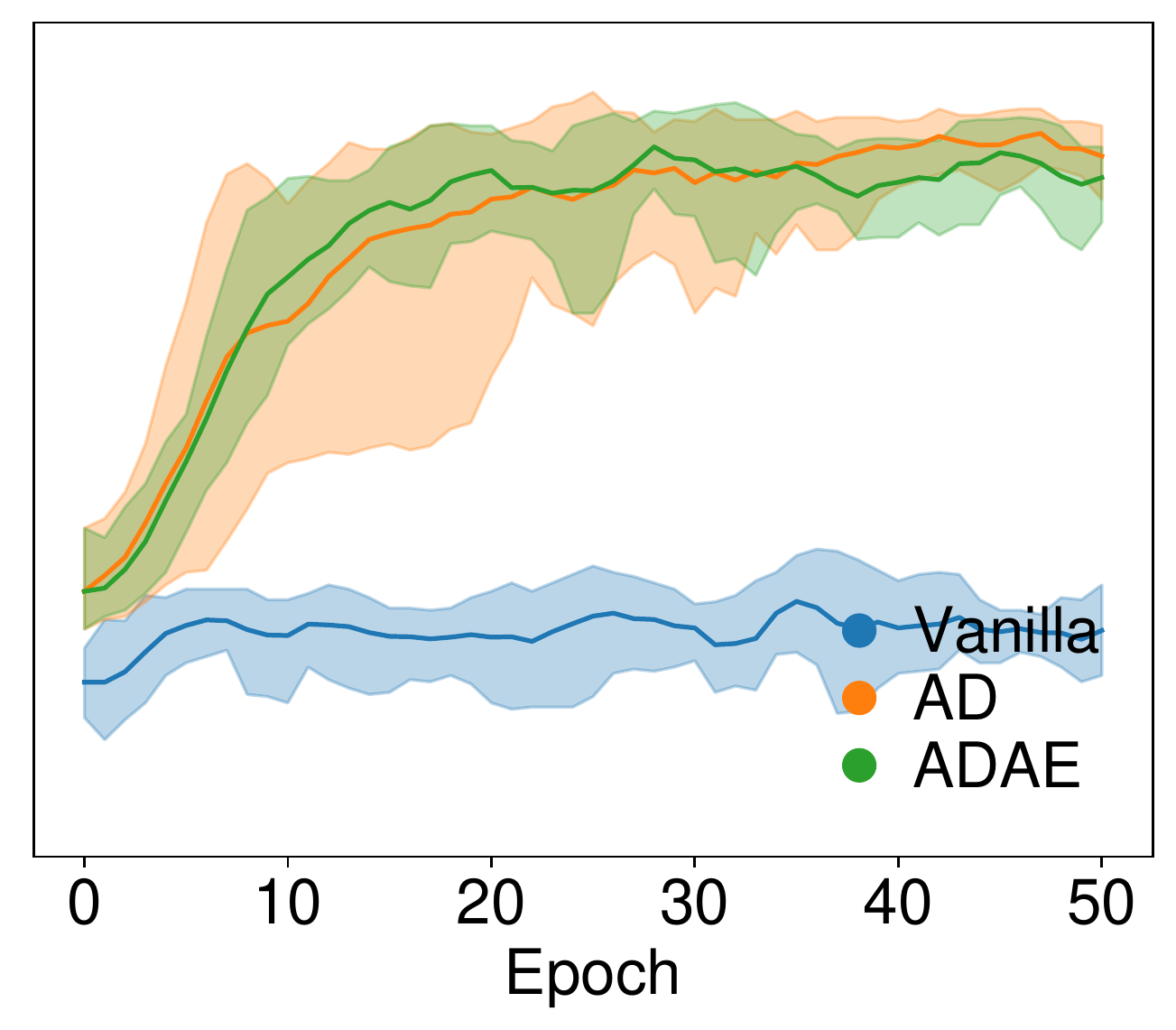}}
	\hfil
	\subfloat[\label{subfig:sac-bs}BlockStack]{
	\includegraphics[height=0.126\textwidth]{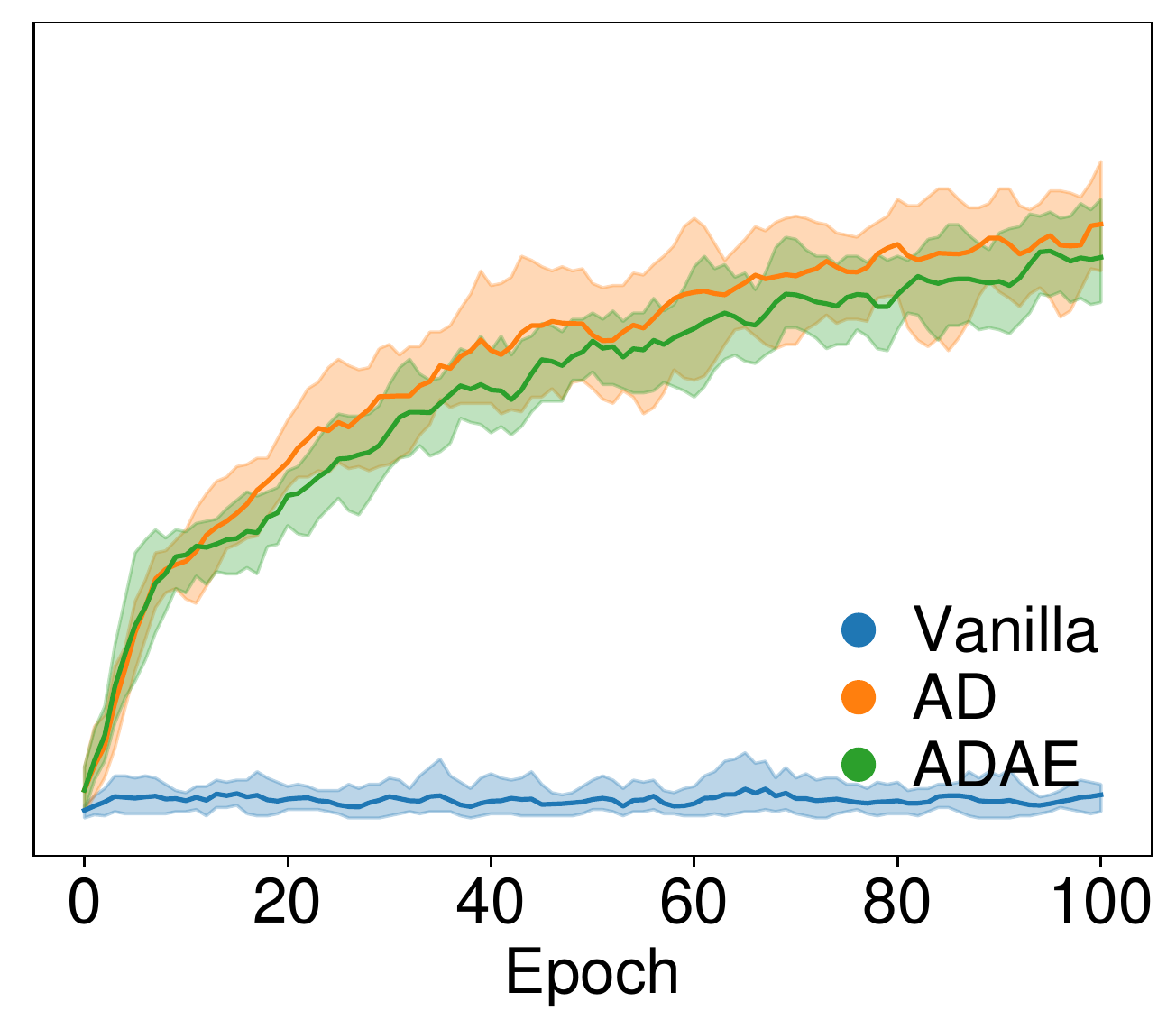}}
	
	\caption{Test success rates of achieving the final goal in all tasks: DQN on gridworld tasks (a-c), DDPG (d-f) and SAC (g-i) on robotic tasks. \textit{AD}: abstract demonstrations; \textit{ADAE}: abstract demonstrations and adaptive exploration.\label{fig:general-performance}}
	\vspace{-15pt}
\end{figure}

As shown by Fig.~\ref{fig:general-performance}, in all experiments, by adding abstract demonstrations in $75\%$ of the training episodes, the agent learns faster with higher performances. The improvement is more obvious in the robotic tasks. Notice that as the robotic task becomes more difficult (from subfigures~\ref{subfig:ddpg-cp} to \ref{subfig:ddpg-bs}), the gap of success rates becomes larger compared to the vanilla algorithm (blue lines). This demonstrates that abstract demonstrations can provide vast improvement on multi-step tasks, because it significantly eases the agent from the difficulty of exploration in long task horizon with sparse rewards and subtask dependencies.

Adaptive exploration provides less obvious improvements on top of abstract demonstrations in terms of success rates. However, it clearly stablises the learning performance, as it shows a smaller variance. This is probably due to its effect on reducing unnecessary exploration such that the agent could act more decisively on well-mastered subtasks.

%===============================================================================

\section{Conclusion}
\label{sec:conclusion}

We introduced $\mathbf{A^2}$ -- abstract demonstration and adaptive exploration -- to aid reinforcement learning algorithms in multi-step, long-horizon and sparse reward tasks. We showed in section~\ref{sec:methods} that $\mathbf{A^2}$ can be integrated in value-based algorithms with discrete actions (e.g., DQN) and actor-critic algorithms with continuous actions with a deterministic (e.g., DDPG) or stochastic (e.g., SAC) policy. We evaluated $\mathbf{A^2}$ in discrete gridworld and continuous robot manipulation environments. Results in section~\ref{sec:result} showed that abstract demonstrations in general speed up learning with higher success rates, with a more significant gain in continuous robot manipulation tasks, and the adaptive exploration module helps the agent to learn more stably.

A limitation of our method is the requirement of a manually designed task decomposition scheme. This could be addressed by learning to decompose a long-horizon task into subtasks. More sophisticated task assumptions such as image observations may also be considered to evaluate the effects of our method.

\printbibliography

\end{document}